# Strategies for Generating Micro Explanations for Bayesian Belief Networks

Peter Sember and Ingrid Zukerman

Department of Computer Science
Monash University
Clayton, VICTORIA 3168, AUSTRALIA
netmail address: pps@bruce.cs.monash.oz@seismo.css.gov
ingrid@bruce.cs.monash.oz@seismo.css.gov

## Abstract

Bayesian Belief Networks have been largely overlooked by Expert Systems practitioners on the grounds that they do not correspond to the human inference mechanism. In this paper, we introduce an explanation mechanism designed to generate intuitive yet probabilistically sound explanations of inferences drawn by a Bayesian Belief Network. In particular, our mechanism accounts for the results obtained due to changes in the causal and the evidential support of a node.

## 1. Introduction

Probability theory or, more accurately, Bayesian Belief Networks (BBNs), is theoretically the soundest formalism for handling uncertainty (Pearl 1986). However, Expert Systems practitioners have largely ignored Bayes Rule on the grounds that it does not correspond to the human inference mechanism (Kahneman and Tversky 1982, Hink and Woods 1987), and have devised new calculi intended to resemble more closely this mechanism (Cohen 1985, Gordon and Shortliffe 1985, Shortliffe 1976). These calculi, however, have been shown to have some undesirable properties (Adams 1976, Buchanan and Shortliffe 1984, Zadeh 1984).

The mechanism presented in this paper seeks to bridge the gap between probabilistic inference and the human interpretation of uncertainty, by generating intuitively appealing and yet probabilistically sound explanations. This approach has been adopted by other researchers (Reggia and Perricone 1985, Norton 1986, Elsaesser 1987, Langlotz, Shortliffe and Fagan 1987). Elsaesser has implemented a system that accounts for special cases that arise in uncertain reasoning. Norton's system is more general, but provides only partial justifications for the obtained results, and Langlotz et al. focus on generating explanations of inferences drawn by decision trees.

A given process or line of reasoning may be explained at different levels of abstraction. The selection of an appropriate level depends on a listener's ability and on the complexity of the process under consideration (Paris 1987). In order to follow the thread of reasoning expressed by a BBN, we consider two basic levels of abstraction: (1) *Macro* and (2) *Micro*.

i.  The Macro level follows the main lines of reasoning leading to a conclusion, without entering into extensive detail. It explains qualitatively the reasoning over an entire network or over a path in the network.

ii. The Micro level presents a detailed account of the transition from one set of beliefs to another in a particular node. It is generated in circumstances in which a high level explanation may be insufficient. This happens in particular when the results obtained at this level don't match human intuition.

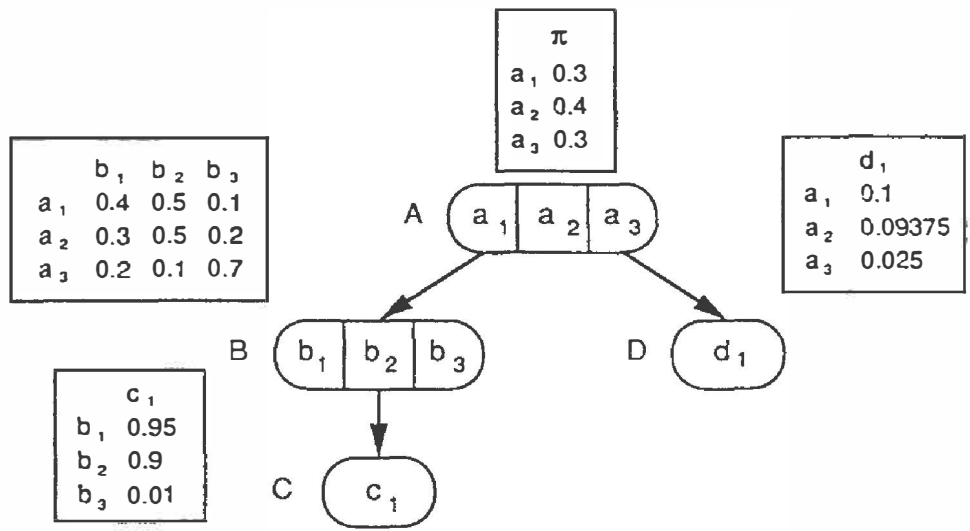

At $t_0$ ($c_1$ and $d_1$ are not grounded):

$\pi(b_1) = 0.30$, $\pi(b_2) = 0.38$, $\pi(b_3) = 0.32$
$\lambda(b_1) = 1.0$, $\lambda(b_2) = 1.0$, $\lambda(b_3) = 1.0$
$Bel(b_1) = 0.30$, $Bel(b_2) = 0.38$, $Bel(b_3) = 0.32$

At $t_1$ ($c_1$ is grounded):

$Bel(a_1) = 0.3955$, $Bel(a_2) = 0.4678$, $Bel(a_3) = 0.1367$
$\pi(b_1) = 0.30$, $\pi(b_2) = 0.38$, $\pi(b_3) = 0.32$
$\lambda(b_1) = 0.95$, $\lambda(b_2) = 0.9$, $\lambda(b_3) = 0.01$
$Bel(b_1) = 0.4522$, $Bel(b_2) = 0.5427$, $Bel(b_3) = 0.0051$

At $t_2$ ($d_1$ is grounded):

$Bel(a_1) = 0.455$, $Bel(a_2) = 0.505$, $Bel(a_3) = 0.04$
$\pi(b_1) = 0.33$, $\pi(b_2) = 0.46$, $\pi(b_3) = 0.21$
$\lambda(b_1) = 0.95$, $\lambda(b_2) = 0.9$, $\lambda(b_3) = 0.01$
$Bel(b_1) = 0.429$, $Bel(b_2) = 0.568$, $Bel(b_3) = 0.003$

Fig. 1: Sample Bayesian Belief Network

The Micro Explanations accounted for by the mechanism presented in this paper focus on a hypothesis of interest in a given node, denoted *focal hypothesis*. The belief in a focal hypothesis $b_f$, e.g., $b_1$ in figure 1, is given by the following formula (Pearl 1986):

$$Belief(b_f) = \alpha\lambda(b_f)\pi(b_f) \tag{1}$$

where $\lambda$ represents evidential support, and $\pi$ — causal support.

One possible strategy for the explanation of $Bel(b_f)$ considers only the current values of $\pi$ and $\lambda$, yielding explanations such as "We have substantial belief in $b_f$ because it has substantial causal support and high evidential support." This strategy may be suitable for situations where the item of interest is the absolute value of a belief in a hypothesis, without taking into consideration the previous value of this belief, or the values of beliefs in competing hypotheses. However, the explanations produced by this strategy leave something to be desired, since it is more often the case that the item of interest is the value of a belief resulting from changes in the network, relative to its previous value and to competing beliefs.

As is evident from Eq. (1), a change in the belief in a proposition may be caused by a change in its causal support, $\pi$, or its evidential support, $\lambda$. Therefore, it is convenient to view a Micro Explanation as having three basic components: (1) Explanation of a change in causal support, (2) Explanation of a change in evidential support, and (3) Explanation of the result obtained by merging these two measures of belief.

In this paper, we focus on a mechanism for the generation of the last of these components. We present a strategy for the generation of explanations for special cases, such as binary nodes and nodes with three hypotheses, and examine how this strategy extends to multi-valued nodes. Strategies for the generation of the other components appear in [Sember and Zukerman 1989a].

## 2. Mathematical Basis of an Explanation

Using Eq. (1) we can show that if the evidential support for a node remains fixed but the causal support changes, then the direction of a change in $Bel(b_f)$ is given by:

> If $U > 0$ then $Bel(b_f)$ increases.
> If $U < 0$ then $Bel(b_f)$ decreases.
> If $U = 0$ then $Bel(b_f)$ does not change.

where:

$$U = \sum_{i=1}^{n} \lambda(b_i) \left\{ \pi(b_f)_{t_i} \pi(b_i)_{t_1} - \pi(b_f)_{t_i} \pi(b_i)_{t_2} \right\} \qquad (2)$$

$$= \sum_{i=1}^{n} \lambda(b_i) U_{i,f}$$

Similarly, if the causal support remains fixed but the evidential support changes, from Eq. (1) we obtain:

> If $D > 0$ then $Bel(b_f)$ increases.
> If $D < 0$ then $Bel(b_f)$ decreases.
> If $D = 0$ then $Bel(b_f)$ does not change.

where:

$$D = \sum_{i=1}^{n} \pi(b_i) \left\{ \lambda(b_f)_{t_i} \lambda(b_i)_{t_1} - \lambda(b_f)_{t_i} \lambda(b_i)_{t_2} \right\} \qquad (3)$$

$$= \sum_{i=1}^{n} \pi(b_i) D_{i,f}$$

$U_{i,f}$ describes the effect on $Bel(b_f)$ of a change in the causal support of $b_i$ and $b_f$, assuming that $b_i$ is the only competitor of $b_f$. $D_{i,f}$ is identical, except that it is concerned with evidential support [†]. From Eq. (2) we can arrive at the following properties of $U_{i,f}$:

> If $U_{i,f} = 0$ then the interaction between $b_i$ and $b_f$ has no effect on $Bel(b_f)$.
> If $U_{i,f} > 0$ then the interaction between $b_i$ and $b_f$ has an upward effect on $Bel(b_f)$.
> If $U_{i,f} < 0$ then the interaction between $b_i$ and $b_f$ has a downward effect on $Bel(b_f)$.

Replacing this formulation of $U_{i,f}$ with the corresponding factor in Eq. (2) yields results consistent with human intuition. For example, $U_{i,f} < 0$ under the following conditions:

> C1 $\Delta\pi(b_f) < 0$ and $\Delta\pi(b_i) > 0$
> C2 $\Delta\pi(b_f) > 0$ and $\Delta\pi(b_i) > 0$, where $\frac{\Delta\pi(b_f)}{\pi(b_f)_{t_1}} \times 100\% < \frac{\Delta\pi(b_i)}{\pi(b_i)_{t_1}} \times 100\%$
> C3 $\Delta\pi(b_f) < 0$ and $\Delta\pi(b_i) < 0$, where $\frac{|\Delta\pi(b_f)|}{\pi(b_f)_{t_1}} \times 100\% > \frac{|\Delta\pi(b_i)|}{\pi(b_i)_{t_1}} \times 100\%$

*C1* represents a situation where the causal support for $b_f$ decreases and the causal support for a competing hypothesis increases, causing the interaction between these hypotheses to have a downward effect on $Bel(b_f)$. In *C2*, the causal support for both hypotheses increases, however the percent-wise increase for $b_f$ is less than for its competitor. *C3* is similar to *C2* but with a decrease in causal support taking place, and the percent-wise decrease in the support for $b_f$ being larger than for $b_i$. The conditions under which the interaction between $b_i$ and $b_f$ has an upward effect or no effect on $Bel(b_f)$ take a similar form.

### 3. Generating Explanations

For an explanation to be convincing, a user's expectations have to be addressed, in particular if they are violated by the results obtained by a system. Failure to do so may cause the user to experience negative affective responses, which in turn may foster unwillingness to accept a system's findings (Zukerman and Pearl 1986). To cater for this requirement, the procedure for generating an explanation for a change in $Bel(b_f)$ performs the following actions:

1. Identify an expectation.

2. If the expectation is met, then generate a basic explanation. Otherwise, generate an explanation based on identifying the cause of the expectation violation.

This procedure is applied both for justifying a change in $Bel(b_f)$ caused by a change in the $\pi(b_i)$-s with the $\lambda(b_i)$-s fixed, and for explaining a change in $Bel(b_f)$ due to a change in the $\lambda(b_i)$-s with the $\pi(b_i)$-s fixed. The application of this procedure in the latter case requires that the $\lambda$-s be normalized. The discussion in the rest of this paper focuses on the former case.

The following table characterizes likely user expectations for $\Delta Bel(b_f)$ in terms of changes in $\pi$.

---

[†] The results obtained for $D_{i,f}$ correspond to those obtained for $U_{i,f}$, and will not be discussed here.

> *E1 If $\pi(b_f)$ increases then the reader expects $Bel(b_f)$ to increase.*
> *E2 If $\pi(b_f)$ decreases then the reader expects $Bel(b_f)$ to decrease.*
> *E3 If $\pi(b_f)$ doesn't change then the reader expects $Bel(b_f)$ not to change.*

### 3.1 Expectation Met — Generating a Basic Explanation

An expectation is met in the following cases: (1) A binary node, (2) All the competing hypotheses have equal evidential support, and (3) The causal support for each of the competing hypotheses either remains unchanged or is changed in a direction opposite to the change in the support of the focal hypothesis, e.g., if the causal support for the focal hypothesis increases, the causal support for the remaining hypotheses either decreases or remains unchanged. In these cases, the premise of the conditional statement corresponding to the met expectation forms the basis of an explanation, and it is unnecessary to go into further details regarding the interaction between the different hypotheses. For example, the explanation we would generate for $\Delta Bel(b_f)$ when expectation $E1$ holds would take the form:

> The belief in $b_f$ has increased due to an increase in its causal support.

### 3.2 Expectation Not Met — Justifying an Expectation Violation

When solving problems, people take advantage of domain knowledge in order to apply the strongest procedure which accomplishes a desired goal (Nilsson 1980). This approach is also taken in Natural Language Processing, where people try to use the most specific terms and phrases which reflect their ideas (Zernik 1987). We follow this paradigm by identifying specific cases which may be better explained by "tailored explanations."

Let us consider a node with three hypotheses. From Eq. (2) it follows that if an expectation with respect to the focal hypothesis is not met, the evidential support for the hypothesis which agrees with the expectation must be lower than the evidential support for the hypothesis which contradicts it (Sember and Zukerman 1989b). We distinguish between two cases which account for competent explanations, depending on the relative strength of the evidential support for the hypothesis which agrees with the expectation.

*Case 1*: The evidential support for this hypothesis does not affect significantly the contradictory result. Hence, it may be discarded, effectively resulting in the node in question becoming a binary node; and

*Case 2*: The evidential support for this hypothesis significantly reduces the effect of the contradicting hypothesis.

Let us now consider an explanation which may be generated for an extreme instance of the first case. In figure 1, we see that the belief in $b_1$ is altered as its causal support changes due to the grounding of $d_1$. The justification for this change is:

> The causal support for $b_1$ increased by 10%, and the support for $b_2$ increased by over 20%. Now, since there is overwhelming evidence against $b_3$, $b_2$ and $b_1$ remain the only two alternatives, thus they compete against each other. As a result, the overall belief in $b_1$ must decrease.[†]

We can obtain an example of the second case by changing the values of the $\lambda$-s at $t_1$ and $t_2$ to $\lambda(b_1) = 0.2268$, $\lambda(b_2) = 0.7524$ and $\lambda(b_3) = 0.2225$, resulting in the set of beliefs at $t_1$, $Bel(b_1) = 0.16$, $Bel(b_2) = 0.6725$ and $Bel(b_3) = 0.1675$, changing to $Bel(b_1) = 0.16$, $Bel(b_2) = 0.74$ and $Bel(b_3) = 0.1$ at $t_2$. The explanation in this case takes the following form:

> Since the evidential support for $b_3$ is lower than for $b_2$, let us assume for a moment that the evidence rules out $b_3$, thereby bringing $b_2$ into closer competition with $b_1$.
>
> If $b_3$ is ruled out, the fact that the causal support for $b_2$ increases by a larger percentage than for $b_1$ leads to the belief in $b_1$ being reduced.
>
> Now, the decrease in the causal support for $b_3$ has the opposite effect on the belief in $b_1$. Hence, since the evidence doesn't completely rule out $b_3$, it actually diminishes the effect of $b_2$. This explains why the belief in $b_1$ remains fixed.

This explanation is based on the following actions:

1. Temporarily eliminate the competing hypothesis $b_i$ with the lowest evidential support. This hypothesis matches the expectation with respect to $b_f$, i.e., it satisfies the condition $sign(U_{i,f}) \neq sign(\Delta Bel(b_f))$.

2. Explain the effect of the remaining hypothesis.

3. Reinstate the eliminated hypothesis, thereby diminishing this effect.

Note that the second step actually generates an explanation similar to the one presented for Case 1. In our example, the true behaviour for $Bel(b_1)$, i.e., no change, lies between the reader's expectation and the extreme result obtained by ruling out $b_3$. The later part of the explanation, which states how the extreme result is moderated by the reinstatement of $b_3$, accounts for this behaviour.

### 3.2.1 Expectation Not Met — Generalizing the Explanation Strategy

The procedure for generating an explanation for a node with an arbitrary number of hypotheses takes the following form:

1. Choose an *Elimination Threshold (ET)*.

2. Form two sets of hypotheses:

   *In* — contains the hypotheses $b_i$ for which $sign(U_{i,f}) = sign(\Delta Bel(b_f))$ and $\lambda(b_i) > ET$. These hypotheses have relatively high evidential support, and their effect contradicts the expectation with respect to $b_f$.

   *Out* — contains the hypotheses $b_i$ for which $sign(U_{i,f}) \neq sign(\Delta Bel(b_f))$ and $\lambda(b_i) < ET$. These hypotheses have relatively low evidential support, and they agree with the expectation with respect to $b_f$.

3. If the ET has a low value, then generate an explanation based only on the hypotheses in the In

---

[†] Adapted from an explanation provided by J. Pearl (1988).

set. Otherwise, temporarily discard the Out set, generate an explanation based on the In set, and then reinstate the Out set.

This explanation strategy constitutes a generalization of the strategy proposed for a node with three hypotheses. It is based on the rationale that in order to justify why an expectation is not met, we must show that one or more hypotheses which support the expectation are eliminated due to lack of evidential support. These are the hypotheses in the set Out. Notice, however, that for explanations generated by this strategy to be acceptable, the elimination threshold must be selected so that there are no hypotheses $b_i$ such that $sign(U_{i,f}) = sign(\Delta Bel(b_f))$ and $\lambda(b_i) < ET$, i.e., there are no hypotheses which contradict the expectation and whose evidential support lies below the ET.

The elimination threshold is determined by applying heuristics which take into consideration the evidential support for each of the competitors of $b_f$. A threshold is guaranteed to exist, because it can been shown that when the expectation is not met, In and Out contain at least one hypothesis (Sember and Zukerman 1989b). In the case where the ET has a low value, the effect of the hypotheses in Out on $Bel(b_f)$ is negligible, and therefore does not need to be mentioned. As the ET increases, the effect of these hypotheses may need to be discussed.

This strategy supports the generation of an explanation of the following general form for a case where the ET is low, and $\pi(b_f)$ increases but $Bel(b_f)$ decreases.

> The evidence is ruling out the hypotheses in *Out*, bringing $b_f$ into closer competition with the hypotheses in *In*. Since the causal support for the hypotheses in *In* increases by a greater percentage than that of $b_f$, $Bel(b_f)$ decreases.

The explanations generated by this strategy are *concocted* (Wick and Thompson 1989, Chandrasekaran, Tanner and Josephson 1989), in the sense that they address one aspect which "explains away" an observed behaviour, ignoring other events which may have bearing on this behaviour, such as the interactions arising from the hypotheses outside the sets In and Out. Still, these explanations appear to be satisfactory to users with diverse probabilistic backgrounds. In addition, explanations which examine all the hypotheses would be considerably longer than the ones produced by our strategy, and, hence, would be unsuitable for a network where a number of nodes have to be considered.

Our explanation strategy must be extensively tested to fine tune the heuristics for choosing the elimination threshold, and also to ascertain the effect on $Bel(b_f)$ of the changes in the causal support for the hypotheses outside the sets In and Out.

### 4. Conclusion

The mechanism discussed in this paper accounts for intuitively appealing and yet probabilistically sound explanations for BBNs. At present, our mechanism is particularly suited for producing explanations involving a few nodes. In particular, it accounts for the results obtained from a combination of the causal and evidential support of a node. Our mechanism is in the initial stages of its implementation, and work is progressing on a module which produces frame-based output.

The next stage of our work involves the generalization of our mechanism to larger networks. This requires the development of pruning techniques which take into account both salience considerations and user attributes.